# An Effective Method of Image Retrieval using Image Mining Techniques

A.Kannan[1]   Dr.V.Mohan[2]   Dr.N.Anbazhagan[3]

[1]Associate Professor, Department of Computer Applications, K.L.N. College of Engineering,

Sivagangai District, Tamilnadu, India – 630611
kannamca@yahoo.com

[2]Professor & Head, Department of Mathematics, Thiagarajar College of Engineering,
Madurai,Tamilnadu,India - 625 015
vmohan@tce.edu

[3]Reader, Department of Maths, Alagappa University, Karaikudi-04, Tamilnadu,India,
anbazhagan_n@yahoo.co.in

## ABSTRACT

*The present research scholars are having keen interest in doing their research activities in the area of Data mining all over the world. Especially, [13]Mining Image data is the one of the essential features in this present scenario since image data plays vital role in every aspect of the system such as business for marketing, hospital for surgery, engineering for construction, Web for publication and so on. The other area in the Image mining system is the Content-Based Image Retrieval (CBIR) which performs retrieval based on the similarity defined in terms of extracted features with more objectiveness. The drawback in CBIR is the features of the query image alone are considered. Hence, a new technique called Image retrieval based on optimum clusters is proposed for improving user interaction with image retrieval systems by fully exploiting the similarity information. The index is created by describing the images according to their color characteristics, with compact feature vectors, that represent typical color distributions [12].*

## 1.0 INTRODUCTION

In this present scenario, image plays vital role in every aspect of business such as business images, satellite images, medical images and so on. If we analysis these data, which can reveal useful information to the human users. But, unfortunately there are certain difficulties to gather those data in a right way [1]. Due to incomplete data, the information gathered is not processed further for any conclusion.  In another end, Image retrieval is the fast growing and challenging research area with regard to both still and moving images. Many Content Based Image Retrieval (CBIR) system prototypes have been proposed and few are used as commercial systems. CBIR aims at searching image databases for specific images that are similar to a given query image. It also focuses at developing new techniques that support effective searching and browsing of large digital image libraries based on automatically derived imagery features. It is a rapidly expanding research area situated at the intersection of databases, information retrieval, and computer vision. Although CBIR is still immature, there has been abundance of prior work.

The CBIR focuses on Image 'features' to enable the query and have been the recent focus of studies of image databases. The features further can be classified as low-level and high-level features. Users can query example images based on these features such as texture, colour, shape, region and others. By similarity comparison the target image from the image repository is retrieved. Meanwhile, the next important phase today is focused on clustering techniques. Clustering algorithms can offer superior organization of multidimensional data for effective retrieval. Clustering algorithms allow a nearest-neighbour search to be efficiently performed.

Hence, the image mining is rapidly gaining more attention among the researchers in the field of data mining, information retrieval and multimedia databases. Image mining presents special characteristics due to the richness of the data that an image can show. Effective evaluation of the results of image mining by content requires that the user point of view (of likeness) is used on the performance parameters [14].

## 1.1 Comparison of Image Mining with other Techniques

The researches in image mining can be classified into two kinds. The image processing is one in which, it involves a domain specific application where the focus is in the process of extracting the most relevant image features into a suitable form[2,3,4] and the image mining is one in which, it involves general application where the focus is on the process of generating image patterns that may be helpful in the understanding of the interaction between high-level human perception of images and low-level features [5,6]. So, the latter may be the best one to lead the improvement in the accuracy of images retrieved from image databases.

Image mining normally deals with the extraction of implicit knowledge, image data relationship, or other patters not explicitly stored from the low-level computer vision and image processing techniques. i.e.) the focus of image mining is the in the extraction of patterns from a large collection of images, the focus of computer vision and image processing techniques is in understanding or extracting specific features from a single image.

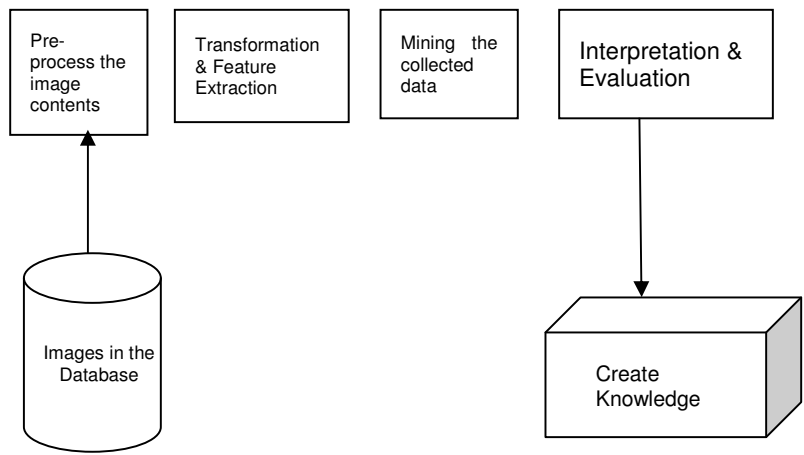

Figure 1 Image Mining Processes [1]

Figure 1.1 shows the image mining process. The images from an image database are first preprocessed to improve their quality. These images then undergo various transformations and feature extraction to generate the important features from the images. With the generated features, mining can be carried out using data mining techniques to discover significant patterns. The resulting patterns are evaluated and interpreted to obtain the final knowledge, which can be applied to applications. [1]

Current techniques in image retrieval and classification (two of the dominant tasks in Image Mining) concentrate on content-based techniques [RHC99]. Various systems like the QBIC [NB94], RetrievalWare [D93] and PhotoBook [PPS96] etc have a variety of features, but are still used in particular domains. Jain et al [JV96] use color features combined with shape for classification. Ma et al [MDM97] use color and texture for retrieval. Smith and Chang [SC96] use color and the spatial arrangements of these color regions. Since perception is subjective, there is no single feature which is sufficient [RHC99, ZHL001]; and, moreover, a single representation of a feature is also not sufficient. Hence multiple representations and a combination of features are necessary [15].

## 2.0 PROBLEM DEFINITION

In the colour based image retrieval the RGB Colour model is used. Colour images normally are in three dimensional. RGB colour components are taken from each and every image. Then the average value of R, G, and B values for both query image and target images are calculated. These three average values for each image are stored and considered as features. By using these stored features the target image from the repository is retrieved with respect to the query image.

Then the top ranked images are re-grouped according to their texture features. In the texture-based approach the parameters gathered are on the basis of statistical approach. Statistical features of grey levels were one of the efficient methods to classify texture. The Grey Level Co-occurrence Matrix (GLCM) is used to extract second order statistics from an image. GLCMs have been used very successfully for texture calculations [9]. The different texture parameters like entropy, contrast, dissimilarity, homogeneity, standard deviation, mean, and variance of both query image and target images are calculated. From the calculated values the required image from the repository is extracted.

Then, the pre-processed images in the database are classified as low-texture, average-texture and high-texture detailed images respectively on the basis of some factor like MLE (Maximum Likelihood Estimation) estimation. The classified images are then subjected to colour feature extraction. The retrieved result is pre-clustered by Fuzzy-C means technique. This is followed by GLCM texture parameter extraction where the texture factors like contrast, correlation, mean, variance and standard variance are mined. The resulted values of both the query image and target images are compared by Euclidean distance method.

### 2.1 Proposed Solution
In this, a new method for image classification is formulated in order to reduce the searching time of images from the image database. The coarse content of image is grouped under three categories as:

(i) High-texture detailed Image
(ii) Average-texture detailed Image
(iii) Low-texture detailed Image

Thereby, we can reduce the search space by one third of what was earlier. If we go more number of groups or less number of groups, they may reveal unnecessary overlapping overhead problems or may produce approximate results. So, the main focus on this classification is by making use of "textures" present in an image. This is because this texture-based classification is simple, easy and efficient for real time applications as compared to classifications based on Entropy method as well as segmentation based techniques.

The primary objective of this work is to develop algorithms in order to create true databases from collections of multimedia data (specifically images) by mining content from the data off-line in order to efficiently support complex queries at run-time [16].

## 2.2 Image Retrieval

Image Retrieval from the image collections involved with the following steps
- Pre-processing
- Image Classification based on some true factor
- RGB processing
- Preclustering
- Texture feature extraction
- Similarity comparison
- Target image selection

## 2.3 BLOCK DIAGRAM: IMAGE RETRIEVAL SYSTEM

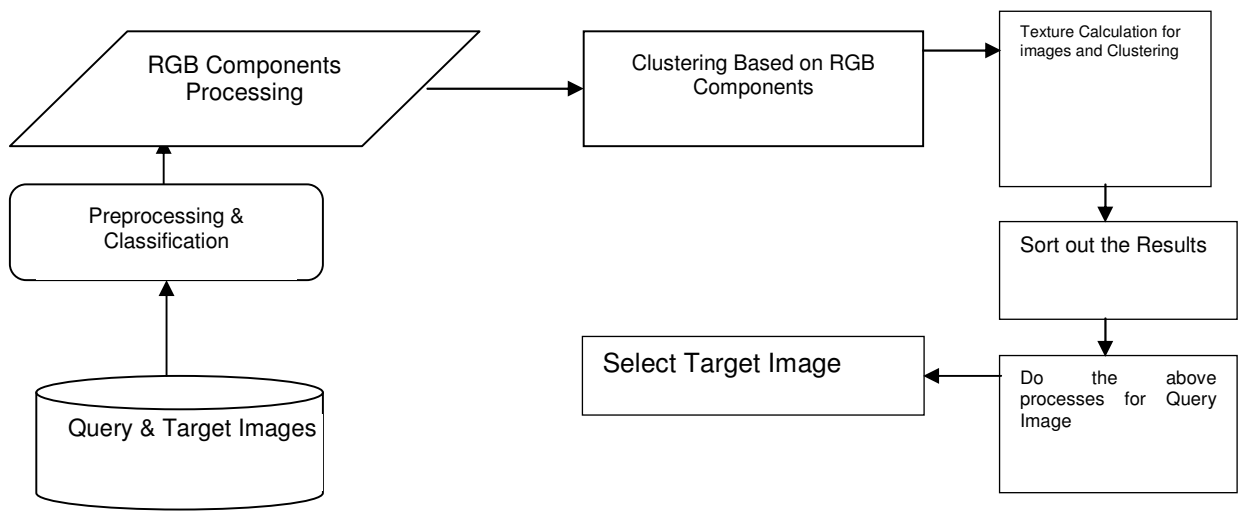

**Figure 2.0**

## 2.4 Pre-processing

Pre-processing is the name used for operations on images at the lowest level of abstraction. The aim of the pre-processing is an improvement of the image that suppresses unwilling distortions or enhances some image features, which is important for future processing of the images. This step focuses on image feature processing.

## 2.5 Noise Reduction Filtering

Filtering is a technique for modifying or enhancing an image. The image is filtered to emphasize certain features or remove other features. The noise in the images is filtered using linear and non-linear filtering techniques. Median filtering is used here to reduce the noise.
Steps for median filtering

- Read the image and display it.
- Add noise to it.
- Filter the noisy image with an averaging filter and display the results.
- Now use a median filter to filter the noisy image and display the results.

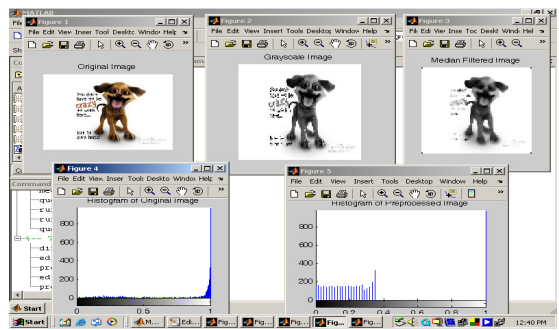

*Figure 2.1 Results for Pre-processing Image*

## 2.6 RGB Components Processing

An RGB colour images is an M*N*3 array of colour pixels, where each colour pixel is a triplet corresponding to the red, green, and blue components of an image at a spatial location. An RGB image can be viewed as the stack of three gray scale images that, when fed into the red, green, blue inputs of a colour monitor, produce the colour image on the screen. By convention the three images form an RGB images are called as red, green and blue components.

The average values for the RGB components are calculated for all images

$$\text{Red average} = \frac{\text{sum of all the Red Pixels in the image R (P)}}{\text{No. Of pixels in the image P}}$$

$$\text{Green average} = \frac{\text{sum of all the Green Pixels in the image G (P)}}{\text{No. Of pixels in the image P}}$$

21

B average= sum of all the Blue Pixels in the image B (P)
              No. Of pixels in the image P

Where R (P) = RED component pixels,
G (P) = GREEN component pixels,
B (P) = BLUE component pixels,
P =No. of pixels in the image.

After calculating the mean values of Red, Blue and Green components, the values are to be compared with each other in order to find the maximum value of the components. For eg., if the value of Red component is High than the rest of the two, then we can conclude that the respective image is Red Intensity oriented image and which can be clustered into Red Group of Images.
Whenever the query image is given calculate the RGB components average values. Then compare this with the stored values.

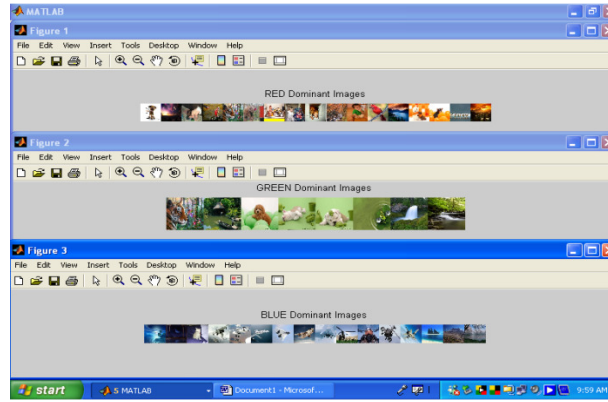

*Figure 2.2 Result of RGB Components Clustering Images*

## 2.7 Image Texture Classification

The texture represents the energy content of the image. If an image contains more and high textures, then the energy will be high as compared to that of average and low texture images. So when combining the energy values defined for a local patch of an image the values will be high for highly textured areas and will be low for smooth areas. Also the local patches of same kind of textured areas will approximate same energy level. So it can be effectively called the "Texture Activity Index". If it is tried to fit the energy values in to any distribution then the classification of images into High, Average and Low detail images can be easily and effectively done because the statistical parameters of the respective distribution will be different for all the three categories as because they possess different energy levels.
The calculated MLE value varies for all the three kind of images. The boundaries for the three categories are fixed based on experimental values. The following fig-3 shows the result of texture analysis calculation.

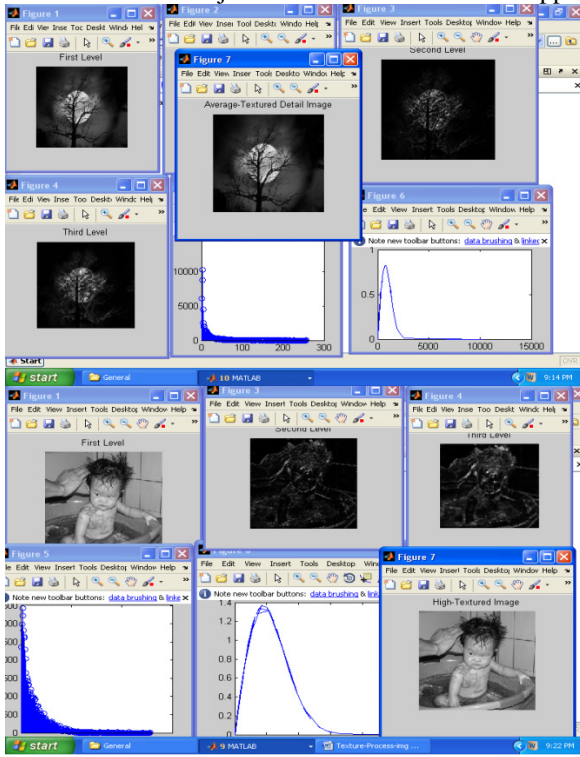

*Figure 2.3 Processes for High and Average Texture Analysis*

## 2.8 Image Clustering

Clustering will be more advantage for reducing the searching time of images in the database. Fuzzy C-means (FCM) is one of the clustering methods which allow one piece of data to belong to two or more clusters. In this clustering, each point has a degree of belonging to clusters, as in fuzzy logic, rather than belonging completely to just one cluster. Thus, points on the edge of a cluster may be in the cluster to a lesser degree than points in the centre of cluster. FCM groups data in specific number of clusters.

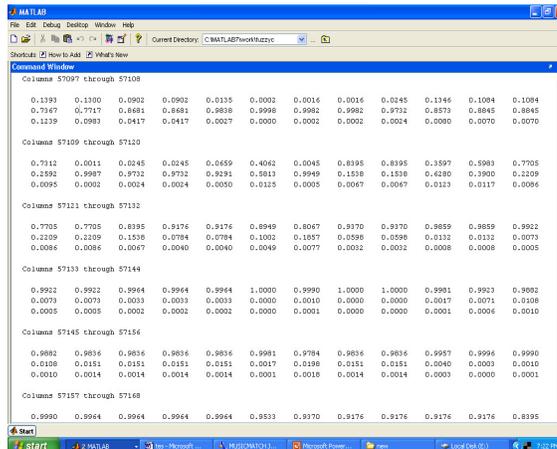

*Figure 2.4 Results of Clustering Processes*

## 2.9 Similarity Comparison

The retrieval process starts with feature extraction for a query image. The features for target images (images in the database) are usually precomputed and stored as feature files. Using these features together with an image similarity measure, the resemblance between the query image and target images are evaluated and sorted. Similarity measure quantifies the resemblance in contents between a pair of images. Depending on the type of features, the formulation of the similarity measure varies greatly. The Mahalanobis distance and intersection distance are commonly used to compute the difference between two histograms with the same number of bins. When the number of bins is different, the Earthmover's distance (EMD) is applied. Here the Euclidean distance is used for similarity comparison.

## 2.10 Neighbouring Target Image Selection

Collections of target images that are "close" to the query image are selected as the neighbourhood of the query image. The major difference between a cluster-based image retrieval system and CBIR systems lies in the two processing stages, selecting neighbouring target images and image clustering, which are the major components of this image retrieval system. A typical CBIR system bypasses these two stages and directly outputs the sorted results to the display and feedback stage. This system can be designed independent of the rest of the components because the only information needed for the system is the sorted similarities. This implies that this module may be embedded in a typical CBIR system regardless of the image features being used, the sorting method, and whether there is feedback or not. The only requirement is a real-valued similarity measure satisfying the symmetry property.

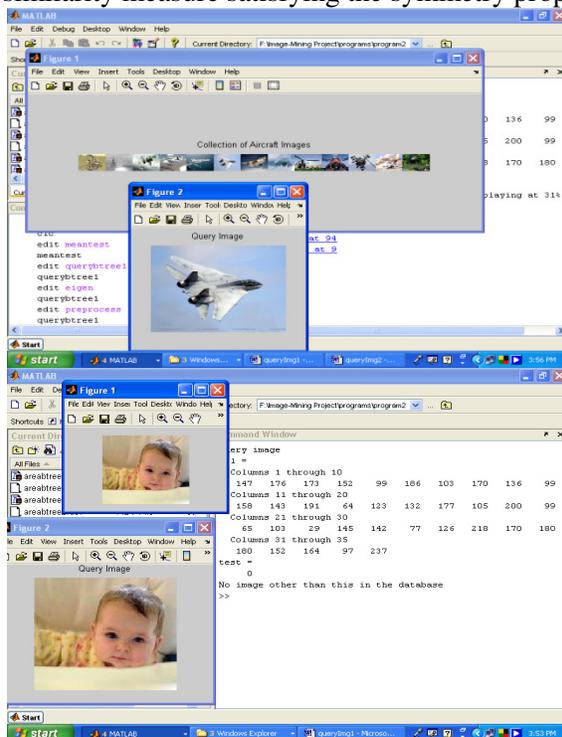

*Figure 2.5 Results of Image Retrieval for the given Query Image*

The International journal of Multimedia & Its Applications (IJMA) Vol.2, No.4, November 2010## 3.0 Performance Evaluation of Proposed CBIR System

Evaluation of retrieval performance is a crucial problem in Content-Based Image Retrieval (CBIR). Many different methods for measuring the performance of a system have been created and used by researchers. We have used the most common evaluation methods namely, Precision and Recall usually presented as a Precision vs Recall graph. Precision and recall alone contain insufficient information. We can always make recall value 1 just by retrieving all images. In a similar way precision value can be kept in a higher value by retrieving only few images or precision and recall should either be used together or the number of images retrieved should be specified.

With this, the following formulae are used for finding Precision and Recall values.

$$\text{Precision} = \frac{No.\ of\ relevant\ images\ retrieved}{Total\ no.\ of\ images\ retrieved} \qquad \text{Recall} = \frac{No.\ of\ relevant\ images\ retrieved}{Total\ no.\ of\ relevant\ images\ in\ the\ database}$$

## 4.0 CONCLUSION

The main objective of the image mining is to remove the data loss and extracting the meaningful potential information to the human expected needs. There are several Content Based Image Retrieval Systems existing in this present scenario. However, this particular system will be applied in Medical transcription in an effective manner not only based on the contents of the image but based on the given query image too for comparing certain frequent diseases affected earlier in human bodies. In this system, a new image retrieval technique based on clusters is also introduced in order to reduce the searching time space. Moreover, the RGB components of the colour images are classified in different dimension in order to create Red, Blue and Green image clusters.